# Comprehensive Evaluation of Large Language Model Responses: A Multi-Factor Scoring System

Yiming Gai[12][0009-0002-0947-7182], Junde Lu[12][0009-0003-3072-6195], Xuefei Huang[2(✉)][0000−0002−8670−1283] and Ying Li[12(✉)] [0000-0002-1545-1811]

[1] School of Computer Science and Engineering, Beihang University, Beijing, China
**{gaiym,ljd2406107,liying}@buaa.edu.cn**
[2] Data Science and Intelligent Computing Laboratory,
Hangzhou International Innovation Institute, Beihang University, Hangzhou, Zhejiang 311115, P.R.China
**{xuefei.huang}@buaa.edu.cn**

**Abstract.** The remarkable performance of large language models (LLMs) in linguistic tasks underscores an urgent need for comprehensive evaluation of their response quality. Prevailing methods, often confined to singular dimensions, fall short of capturing the full spectrum of model capabilities. This study introduces a multifactor scoring paradigm, integrating accuracy, conciseness, factual consistency, readability, and coherence, complemented by a graphical user interface (GUI) for visualizing outcomes. Evaluations on the TruthfulQA dataset unveil mainstream LLMs' strengths in reasoning tasks (peaking at a composite score of 0.6104) alongside pervasive limitations in navigating complex facts and ambiguities. Transcending the narrow lens of traditional metrics, this framework offers a transparent, adaptable avenue to illuminate model potential and deficiencies. Though presently focused on English tasks, its horizons beckon toward multilingual domains. This work carves a novel path for knowledge engineering and model refinement.

**Keywords:** LLM Evaluation, Multi-factor Scoring, Large Language Models, Benchmarking, Model Comparison

## 1 Introduction

LLMs have permeated various aspects of human life, from intelligent assistants to academic research, and their influence is ubiquitous. However, a scientific and comprehensive evaluation of LLM's strengths and weaknesses, particularly its performance across multiple testing criteria, remains a pressing challenge. Traditional evaluation methods have primarily focused on a single dimension, such as accuracy or fluency. However, with the diversification of application scenarios, relying on a single metric is no longer sufficient to fully capture the model's capabilities. As a result, the development of multi-dimensional evaluation frameworks has become a key research focus.

Existing research has proposed various methods for evaluating the quality of LLM responses. Metrics based on n-gram matching, such as BLEU and ROUGE, are widely used in machine translation and text generation tasks[1]. However, their limitations in semantic understanding and contextual coherence make them less suitable for open-domain question answering. Recently, semantic embedding-based evaluation methods have gained traction. BERTScore[2], for example, leverages pre-trained models to compute semantic similarity between texts, addressing the shortcomings of traditional metrics. For factual consistency, datasets like TruthfulQA[3] have been developed to test a model's truthfulness and reasoning ability with specially designed questions. However, these methods often overlook user experience-related dimensions such as readability and conciseness, making it difficult to provide a comprehensive evaluation perspective.

In specialized domains, such as healthcare or law, the evaluation requirements for LLMs become more complex. These domains not only demand high factual accuracy but also necessitate the proper use of technical terminology and logical coherence[4]. However, the applicability of existing evaluation methods has not been fully validated in these contexts, as traditional metrics and single-dimensional assessments are inadequate for addressing the diverse practical needs. To address this, this paper proposes a multi-factor scoring system that combines five key dimensions: accuracy, conciseness, factual consistency, readability, and coherence, while also integrating the ROUGE metric, as shown in the Fig.1. The aim is to provide a comprehensive and scientific evaluation framework for LLM response quality, validated through the TruthfulQA dataset. Experimental results demonstrate that this framework effectively highlights LLMs' strengths in reasoning tasks and their pervasive limitations in handling complex factual scenarios.

The contributions of this paper are as follows:

***A. Proposing a multi-dimensional evaluation framework***. This study innovatively designs a comprehensive scoring system that incorporates traditional single metrics (such as accuracy or fluency). By incorporating user experience-related dimensions such as readability, conciseness, and coherence, it provides a holistic characterization of LLM performance across diverse tasks.

***B. Enhancing evaluation applicability and transparency.*** By integrating semantic embedding techniques with the ROUGE metric and validating the approach using the TruthfulQA dataset, this method is not only applicable to open-domain question answering but also provides a scalable solution for the complex evaluation needs of specialized domains (such as healthcare and law). Additionally, it enhances the visualization and interpretability of results through the use of a graphical user interface (GUI).

***C. Revealing model capabilities and limitations.*** Through multi-factor analysis, this study systematically reveals the strengths of LLMs in reasoning tasks (such as Logical Falsehood) and their limitations when handling complex facts and ambiguous information (such as Misinformation). This provides data-driven insights and theoretical guidance for future model optimization.

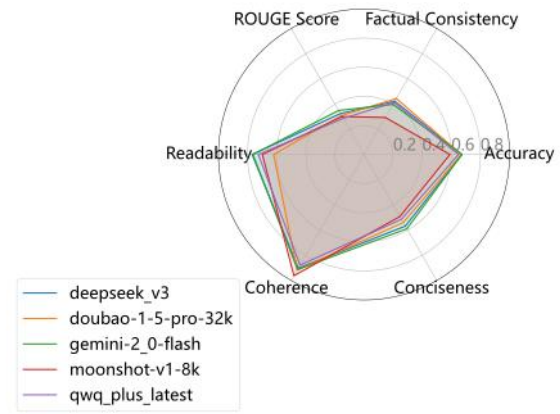

**Fig. 1.** Pie chart of the scores of five LLMs on different dimensions based on TruthfulQA

## 2 Related work

Large language models (LLMs), based on the Transformer architecture and trained on extensive corpora, excel in NLP tasks like text generation and question answering [5]. Deep learning, a key driver in AI, leverages neural networks—modeled on the human brain—to advance pattern recognition, vision, and NLP [6,7]. Neurons process inputs, apply transformations, and adjust outputs via weighted connections [8]. By exploiting inter-layer links, deep learning extracts hierarchical features, enhancing data representation [9]. The Transformer's self-attention mechanism has transformed NLP feature extraction.

Traditional evaluation metrics such as BLEU[1] and ROUGE[12] have historically been employed to assess machine translation and summarization models by measuring n-gram overlap. However, these metrics primarily capture surface-level lexical similarities and fail to account for deeper semantic relationships or contextual coherence — critical components of human language understanding. While effective for syntactic comparisons, these approaches exhibit notable limitations in evaluating generative models' comprehension and reasoning capabilities. To address these shortcomings, a wave of semantic-oriented evaluation methods has emerged. BERTScore[2] leverages pre-trained contextual embeddings to measure text similarity, offering a more nuanced assessment of semantic alignment. TruthfulQA[3] advances this paradigm by designing adversarial questions that probe the factual consistency and reasoning capabilities of LLMs, revealing their robustness in knowledge-grounded scenarios. An essential component underpinning these advancements is tokenization, particularly Byte Pair Encoding (BPE)[10]. BPE iteratively merges frequent character pairs, striking a balance between vocabulary efficiency and linguistic expressiveness. This method significantly enhances model performance in English-centric benchmarks such as TruthfulQA, enabling more precise evaluations of LLM capabilities. Despite these innovations, existing evaluation methods remain constrained to specific dimensions—either focusing on semantic similarity or factual accuracy—thereby offering an incomplete picture of model proficiency.

The limitations of traditional evaluation methods have spurred the development of multidimensional frameworks. Beyond automated metrics, human evaluation offers critical insights into fluency, coherence, and informativeness through direct analysis or comparison with human-authored texts [4]. ROUGE variants, such as ROUGE-N and ROUGE-L, measure content overlap using n-gram matching and longest common subsequences, capturing lexical recall and structural similarity [13]. The F1 score balances precision and recall to assess answer relevance and completeness. However, these metrics struggle to evaluate readability, logical flow, and user perception. Semantic redundancy analysis [11] quantifies repetition but lacks adaptability to dynamic content, while factual consistency assessments, often reliant on keyword matching or external knowledge retrieval [8], fall short in open-domain question answering. Evaluation complexity increases in specialized domains like medicine and law, where precision and coherence are critical [4]. Conventional approaches frequently fail in these contexts, revealing a gap between metric-driven assessment and practical utility. Consequently, multidimensional frameworks have gained traction, integrating diverse perspectives to comprehensively assess model capabilities. Motivated by these challenges, this study proposes a multi-factor scoring system incorporating accuracy, conciseness, factual consistency, readability, and coherence, informed by ROUGE-based evaluation [13]. Leveraging TruthfulQA as a benchmark, our approach transcends single-metric limitations, providing a holistic view of LLMs' strengths and weaknesses in reasoning, factual grounding, and user alignment. This framework offers a structured methodology for model evaluation, optimization, and task-specific adaptation, contributing to advancements in knowledge engineering and AI-driven decision-making.

# 3 Methodology

To comprehensively assess the response quality of large language models (LLMs), we propose a multi-factor scoring system that quantitatively analyzes responses across six dimensions: accuracy, conciseness, factual consistency, readability, coherence, and ROUGE score. This system aims to establish a theoretically rigorous, technologically advanced, and practically viable evaluation framework. This section elaborates on the design principles, core algorithms, metric computation methodologies, and comparisons with existing evaluation approaches.

## 3.1 System Design Concept and Technical Framework

The design of this system is motivated by the theoretical necessity of multi-dimensional evaluation, as a single metric is insufficient to comprehensively capture the diverse performance of LLMs across various tasks[14]. To address this challenge, we adopt a modular architecture comprising a data preprocessing module, a semantic embedding module, and a multi-factor evaluation module, ensuring both methodological flexibility and system scalability.

The data preprocessing module standardizes input text, handles missing values and non-string inputs, and unifies sentence segmentation rules using regular expressions ([.!?]). This step leverages established text normalization techniques[13] to enhance evaluation consistency.

The semantic embedding module serves as the technical core of our system, leveraging pretrained language models to transform textual responses into high-dimensional vector representations. We employ the all-MiniLM-L6-v2 model[15], which is built on the Transformer architecture and offers both computational efficiency and multilingual support. Compared to larger models such as BERT, its lightweight design is more suitable for real-time evaluation scenarios while maintaining robust semantic representation capabilities.

The multi-factor evaluation module computes various assessment metrics based on embedding vectors and integrates them into a composite score using a weighted averaging strategy. This modular framework not only enhances computational efficiency but also facilitates the seamless integration of additional models or evaluation criteria in future iterations.

### 3.2 Theory and Algorithms of Evaluation Metrics

**Accuracy**. Accuracy quantifies the semantic similarity between the model-generated response $A_m$ and the reference answer $A_g$, serving as a fundamental evaluation metric. We employ cosine similarity to measure the alignment between the embedding vectors of the two texts[16]. Given embedding vectors $E_m$ and $E_g$, the accuracy score is defined as:

$$AS = \cos\left(E_m, E_g\right) = \frac{E_m \cdot E_g}{\|E_m\|\|E_g\|}. \tag{1}$$

To ensure robustness, input text undergoes preprocessing (null value checks and string conversions) to prevent invalid computations. Compared to traditional n-gram matching methods (e.g., BLEU), this approach prioritizes semantic consistency over exact word matches.

**Conciseness**. Conciseness evaluates whether the model's response is succinct and free from redundant expressions. Inspired by information redundancy theory[10], we design an inverse redundancy metric. The algorithm first segments $A_m$ and $A_g$ into sentence sequences and then computes each sentence's maximum similarity with the reference. If the similarity falls below a predefined threshold, a full-length penalty is applied; otherwise, the penalty is inversely proportional to the similarity score.For repetitive sentences, an exponential penalty ($2^{\text{repeat_count}}$) is applied, supplemented by a quadratic penalty ($\text{count}^2 \cdot 0.5$). If the response $A_m$ length exceeds twice that of the reference $A_g$, a length-based scaling factor is incorporated. The redundancy score is formulated as:

$$R = \min\left(1.0, \frac{\text{redundancy}}{\text{total length}}\right). \tag{2}$$

The final conciseness score is computed as:

$$CS = 1 - R. \tag{3}$$

**Factual.** Factual consistency assesses the alignment between key factual elements in the model's response and the ground truth. We adopt a set-overlap-based approach[3], wherein both $A_m$ and $A_g$ are tokenized, converted to lowercase, and filtered to remove stopwords (e.g., "the", "and"). Given word sets $W_m$ and $W_g$, the consistency score is calculated as:

$$\mathrm{FC} = \frac{|\mathrm{W_m} \cap \mathrm{W_g}|}{|\mathrm{W_g}|}. \tag{4}$$

If $\mathrm{A_g}$ is empty, the score defaults to 1.0; if either input is empty, the score is 0.0. This approach is computationally efficient and well-suited for English datasets. Future enhancements may incorporate named entity recognition (NER) to improve precision.

**Readability.** Readability measures the ease of comprehension of the generated response, based on linguistic readability theory[17]. We consider two sub-metrics: Sentence Length Score: Deviation from an optimal sentence length of 17.5 words, normalized. Lexical Diversity: The proportion of unique words in the response (+1e-6 smoothing to prevent division by zero).

The overall readability score is defined as:

$$\mathrm{RD} = 0.6 \cdot \left(1 - \min\left(1.0, \frac{|\mathrm{avg_{length}} - 17.5|}{17.5}\right)\right) + 0.4 \cdot \frac{|\mathrm{unique_{words}}|}{|\mathrm{words}| + 10^{-6}}. \tag{5}$$

For single-sentence or empty responses, a default score of 0.5 is assigned to maintain fairness.

**Coherence.** Coherence evaluates the logical flow within the response, grounded in discourse coherence theory[18]. The response $A_m$ is segmented into sentences $S = \{s_1, s_2, \ldots, s_n\}$, and cosine similarity is computed between the embedding vectors of adjacent sentences. The coherence score is obtained by averaging these similarity values:

$$CH = \frac{1}{n-1} \sum_{i=1}^{n-1} \cos\left(E_{s_i}, E_{s_{i+1}}\right). \tag{6}$$

For single-sentence or empty responses, the score defaults to 1.0. This method effectively captures local cohesion, making it particularly suitable for evaluating short-form responses.

**ROUGE Score.** To complement semantic-based evaluations, we incorporate ROUGE scores[12] to quantify n-gram overlap between $A_m$ and $A_g$. Specifically, we compute ROUGE-1, ROUGE-2, and ROUGE-L F1 scores, which are integrated as follows:

$$RS = 0.5 \cdot R_1 + 0.2 \cdot R_2 + 0.3 \cdot R_L. \tag{7}$$

To enhance score differentiation, the final score is scaled by 1.2, with an upper bound of 1.0, thereby emphasizing the importance of ROUGE-1 and ROUGE-L in lexical similarity.

### 3.3 Comprehensive Scoring Method

The final Total Score (TS) is computed using a weighted average of the six evaluation metrics. The default weight assignments are: Accuracy (0.25), Conciseness (0.10), Factual Consistency (0.20), Readability (0.15), Coherence (0.15), and ROUGE Score (0.15). These weights are empirically determined based on task requirements and inter-metric correlations[19]:

$$\mathrm{TS} = \sum_{\mathrm{i}} \mathrm{w_i} \cdot \mathrm{M_i}, \tag{8}$$

where $w_i$ represents the weight and $M_i$ denotes the individual metric scores. This framework allows for dynamic weight adjustments to adapt to different evaluation scenarios.

### 3.4 Comparison with Existing Methods

Compared to traditional evaluation approaches, our system offers several advantages:

BLEU[21]: While BLEU relies on n-gram overlap, our semantic embedding approach enhances semantic sensitivity in evaluation.

BERTScore[16]: Unlike BERTScore, which focuses primarily on semantic similarity, our system introduces conciseness, factual consistency, and user experience metrics (i.e., readability and coherence) to provide a more comprehensive assessment.

ROUGE: By integrating ROUGE, we complement semantic-based evaluation with surface-level text similarity, ensuring a more balanced assessment of both lexical and semantic fidelity.

This multi-factor evaluation framework is theoretically more comprehensive and technically better aligned with the complex application requirements of LLMs.

## 4 Experimental Process and Results

This section describes how we apply the multi-factor scoring system to evaluate the response quality of LLMs, including the experimental setup, data preprocessing, evaluation process, and result analysis. We selected the TruthfulQA dataset to test five mainstream LLMs, and through both quantitative and qualitative analysis, we reveal the performance characteristics of each model.

### 4.1 Experimental Setup

**Dataset Selection.** We selected the TruthfulQA dataset[1], which contains 817 English questions spanning various categories such as science, history, and health, and is designed to test the model's truthfulness and reasoning ability. Each question is paired with an optimal answer, providing a standard reference for evaluation. The choice of TruthfulQA is motivated by its compatibility with the English language and the BPE encoding, which allows LLMs to fully leverage their strengths in English-language tasks.

**Model Selection.** The following five LLMs were evaluated in the experiment:

*qwq_plus_latest*: A multilingual model developed by Alibaba Cloud, known for its efficient reasoning capabilities.

*deepseek_v3*: An open-source model focused on deep learning optimization.

*doubao-1-5-pro-32k*: A conversational model launched by ByteDance, emphasizing generation fluency.

*moonshot-v1-8k*: Known for its strong semantic understanding, instruction following, and text generation abilities.

*gemini-2.0-flash*: A multilingual model that excels in multimodal understanding and reasoning.

**Data Preprocessing.** The experimental data is sourced from the TruthfulQA CSV file, which includes columns such as "Question" and "Best Answer." To ensure consistency in the evaluation, the following preprocessing steps were performed:

*Data Cleaning*. The file was read using pd.read_csv, followed by checks and removal of null values or invalid responses.

*Format Standardization*. Both model responses and reference answers were converted to strings, and non-string inputs were processed to ensure compatibility.

*Metadata Addition*. Model identification (Model), question category (Category), and response type (Type) were added to each record to facilitate grouped analysis.

*File Management*. The responses of each model were stored as separate CSV files, with file paths specified by parameters, and the output results were saved to a designated directory.

### 4.2 Evaluation Process

The evaluation process is implemented based on the evaluate_model_responses function, with the following steps:

1) *Data Traversal*. The data files are read one by one, extracting the questions, model responses, and reference answers.

2) *Metric Calculation*. Six metrics are computed for each pair of responses:

Accuracy : Computes the cosine similarity of the embedding vectors.

Conciseness : Calculates redundancy and takes the inverse.

Factual Consistency : Computes the word set overlap rate.

Readability : Combines sentence length and vocabulary diversity.

Coherence : Computes the average similarity between sentences.

ROUGE Score : Weighted integration of ROUGE-1, ROUGE-2, and ROUGE-L.

**Overall Scoring**. The total score is calculated using the following weights: Accuracy (0.25), Conciseness (0.10), Factual Consistency (0.20), Readability (0.15), Coherence (0.15), and ROUGE Score (0.15).

3) *Result Storage*. A detailed results table and a model summary table are generated, containing the average values for each metric and the total score.

Exception handling is integrated throughout the process. If a file is missing or a calculation error occurs, a warning is recorded and the error is skipped, ensuring the stability of the experiment.

### 4.3 Result Analysis

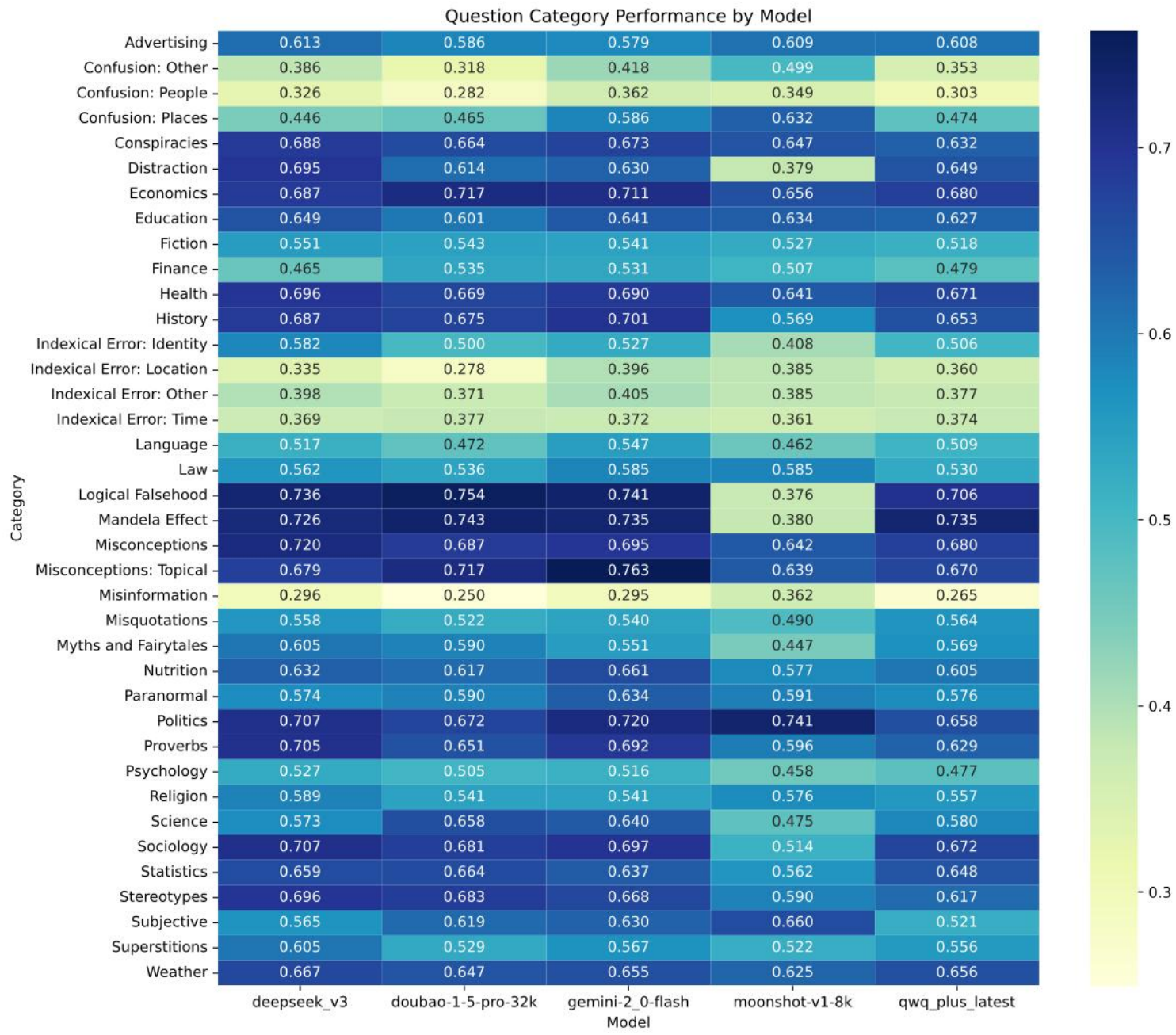

Fig. 2. Category-Wise Performance of LLMs on TruthfulQA

Based on the evaluation results, we can draw the following conclusions, as shown in the Fig.3,Table 1:

**Table 1.** Performance Metrics of LLMs

| Model | Accuracy | Conciseness | Factual Consistency | Total Score |
|---|---|---|---|---|
| DeepSeek-v3 | 0.6719 | 0.5687 | 0.4228 | 0.6088 |
| Doubao-1.5-pro-32k | 0.6689 | 0.5403 | 0.4445 | 0.5873 |
| Gemini-2.0-flash | 0.6725 | 0.5897 | 0.3975 | 0.6104 |
| QWQ Plus Latest | 0.6496 | 0.5145 | 0.4126 | 0.5800 |
| moonshot-v1-8k | 0.5905 | 0.4927 | 0.2936 | 0.5499 |

**Table 2.** Deepseek-V3

| category | score |
|---|---|
| Logical Falsehood | 0.7356 |
| Mandela Effect | 0.7257 |
| Misinformation | 0.2963 |
| Confusion: People | 0.3259 |

DeepSeek-v3 demonstrates an overall performance score of 0.6088, reflecting a mixed capability across various metrics. It excels particularly in handling Logical Falsehood questions, where its strengths shine through, but it struggles notably with Misinformation questions, as shown in the Fig.3 (a),Table 2.

**Table 3.** doubao-1.5-pro-32k

| category | score |
|---|---|
| Logical Falsehood | 0.7540 |
| Mandela Effect | 0.7426 |
| Misinformation | 0.2497 |
| Indexical Error: Location | 0.2784 |

Doubao-1.5-pro-32k exhibits an overall performance score of 0.5873, indicating a varied proficiency across its evaluated metrics. It shines in addressing Logical Falsehood questions, showcasing its strengths in specific reasoning tasks, yet it falters significantly with Misinformation questions, as shown in the Fig.3 (b) ,Table 3.

**Table 4.** gemini-2.0-flash

| category | score |
|---|---|
| Misconceptions: Topical | 0.7628 |
| Logical Falsehood | 0.7414 |
| Misinformation | 0.2950 |
| Confusion: People | 0.3617 |

Gemini 2.0 Flash exhibits an overall performance score of 0.6104, showcasing a balanced yet varied proficiency across its evaluated metrics. It stands out in addressing Misconceptions: Topical questions, where it performs at its peak, but it falters when tackling Misinformation questions, as shown in the Fig.3 (c) ,Table 4.

**Table 5.** qwq_plus_latest

| category | score |
|---|---|
| Mandela Effect | 0.7353 |
| Logical Falsehood | 0.7059 |
| Misinformation | 0.2652 |
| Confusion: People | 0.3027 |

The qwq-plus-latest model achieves an overall performance score of 0.5800, indicating a moderate level of effectiveness across its evaluated metrics. It demonstrates particular strength in handling Mandela Effect questions, where its capabilities stand out, but it shows a relative weakness in addressing Misinformation questions, as shown in the Fig.3 (d) ,Table 5.

**Table 6.** moonshot-v1-8k

| category | score |
|---|---|
| Politics | 0.7406 |
| Subjective | 0.6601 |
| Confusion: People | 0.3492 |
| Indexical Error: Time | 0.3607 |

Moonshot-v1-8k exhibits an overall performance score of 0.5499, indicating a varied proficiency across its evaluated metrics. It demonstrates particular strength in addressing Politics questions, where it likely leverages its capabilities effectively, but it shows notable weakness in handling Confusion: People questions, as shown in the Fig.3 (e) ,Table 6.

Based on the comprehensive analysis of the models' performance across various metrics, we provide the following recommendations tailored to specific scenario requirements. For scenarios where high accuracy is essential, the Gemini 2.0 Flash model is recommended due to its exceptional precision. In cases where factual consistency is the top priority, the Doubao-1.5-pro-32k model stands out as the optimal choice, ensuring reliable and accurate information. Additionally, for situations that demand high readability, the DeepSeek-v3 model is the most suitable option, offering clear and engaging output. These suggestions are designed to address distinct needs effectively, based on the evaluated strengths of each model.

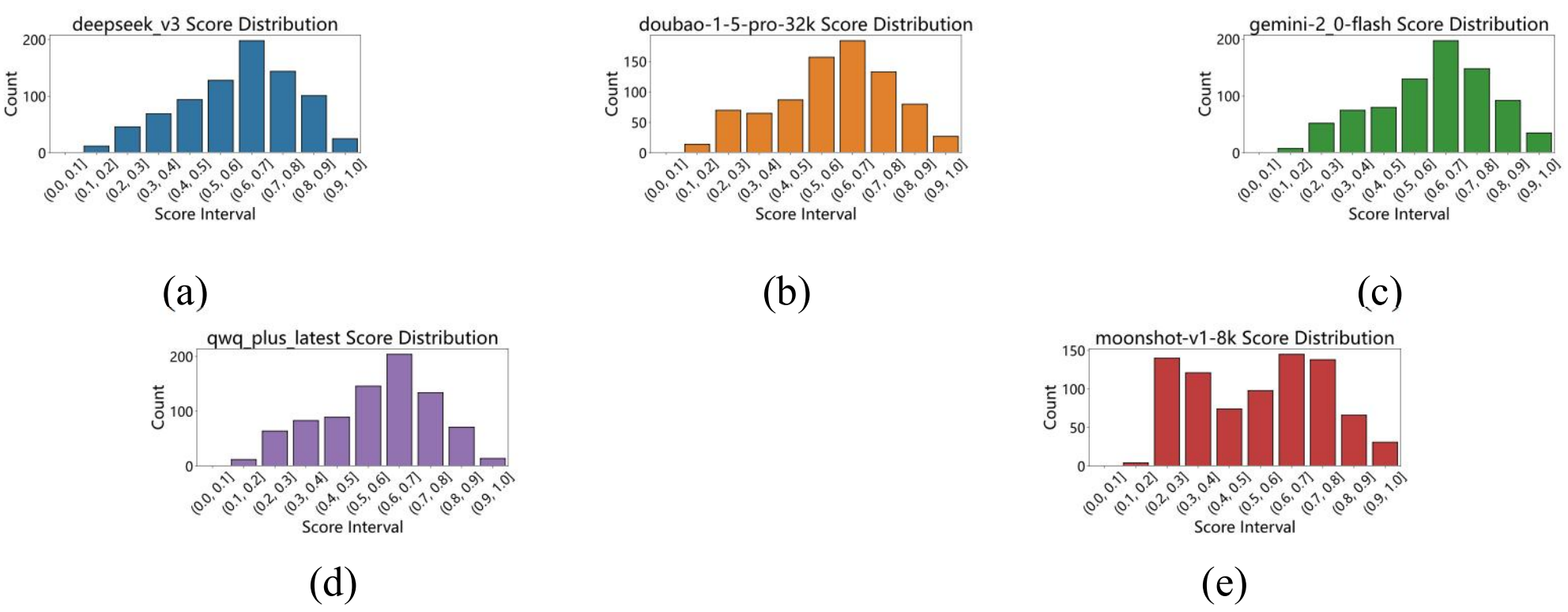

Fig. 3. Performance Metrics of LLMs

## 5 Conclusion

This paper introduces a multi-factor scoring system to assess Large Language Models (LLMs) using the TruthfulQA dataset. It evaluates models on several key metrics: accuracy, conciseness, factual consistency, readability, coherence, and ROUGE scores. Experimental results highlight unique strengths among models: Gemini 2.0 Flash stands out for accuracy and conciseness, DeepSeek V3 for readability and coherence, and Doubao-1.5-Pro-32k for factual consistency. While the models perform well in reasoning, they face challenges with complex facts and ambiguity. Unlike single-metric evaluations, this approach offers a comprehensive framework for model selection and improvement. Currently focused on English, it has potential for future multilingual expansion, multimodal applications, and integration with knowledge graphs and user studies, pending enhancements in factual consistency and weight optimization.